\documentclass[10pt,twocolumn,letterpaper]{article}

\usepackage[pagenumbers]{cvpr} 

\usepackage[dvipsnames]{xcolor}

\usepackage{bm}
\usepackage{amsfonts}
\usepackage{multirow}
\usepackage{makecell}
\usepackage{microtype}
\usepackage[accsupp]{axessibility}

\newcommand{\tbf}[1]{\textbf{#1}}
\newcommand{\tul}[1]{\underline{#1}}

\newcommand{\tworow}[2]{\begin{tabular}[c]{@{}c@{}}#1\vspace{-2pt}\\#2\end{tabular}}

\newcommand{\psp}{\kern0.2ex}
\newcommand{\nsp}{\kern-0.1ex}
\newcommand{\pslp}{\psp/\psp}

\newcommand{\SocialLSTM}{Social-LSTM\,\cite{alahi2016social}}
\newcommand{\SocialGAN}{Social-GAN\,\cite{gupta2018social}}
\newcommand{\SRLSTM}{SR-LSTM\,\cite{zhang2019srlstm}}
\newcommand{\STARD}{STAR-D\,\cite{yu2020spatio}}
\newcommand{\STAR}{STAR\,\cite{yu2020spatio}}
\newcommand{\Trajectronpp}{Trajectron++\psp\cite{salzmann2020trajectron++}}
\newcommand{\SocialVAE}{SocialVAE\,\cite{xu2022socialvae}}
\newcommand{\GroupNet}{GroupNet\,\cite{xu2022groupnet}}
\newcommand{\LED}{LED\,\cite{mao2023leapfrog}}
\newcommand{\EqMotion}{EqMotion\,\cite{xu2023eqmotion}}

\newcommand{\GPGraph}{GP-Graph\,\cite{bae2022gpgraph}}
\newcommand{\NPSN}{NPSN\,\cite{bae2022npsn}}
\newcommand{\LBEBM}{LBEBM\,\cite{pang2021lbebm}}
\newcommand{\STGCNN}{STGCNN\,\cite{mohamed2020social}\!}
\newcommand{\PECNet}{PECNet\,\cite{mangalam2020pecnet}}
\newcommand{\AgentFormer}{A\nsp g\nsp e\nsp n\nsp tF\nsp o\nsp r\nsp m\nsp e\nsp r\,\cite{yuan2021agent}}
\newcommand{\MID}{MID\,\cite{gu2022mid}}
\newcommand{\ETFullCompact}{\!E\nsp i\nsp g\nsp e\nsp n\nsp\nsp\nsp T\nsp\nsp\nsp r\nsp a\nsp j\nsp e\nsp c\nsp t\nsp o\nsp r\nsp y\psp\cite{bae2023eigentrajectory}\!}
\newcommand{\STFullCompact}{\!S\nsp i\nsp n\nsp g\nsp u\nsp l\nsp a\nsp r\nsp\nsp\nsp\nsp\nsp   T\nsp\nsp\nsp  r\nsp a\nsp\nsp j\nsp e\nsp c\nsp t\nsp o\nsp r\nsp y\!}
\newcommand{\TGNN}{T-GNN\,\cite{xu2022tgnn}}
\newcommand{\Kzero}{K0\,\cite{ivanovic2023expanding}}
\newcommand{\Adaptive}{Adaptive\,\cite{ivanovic2023expanding}}

\newcommand{\STT}{STT\,\cite{monti2022stt}}
\newcommand{\STTDTO}{STT+DTO\,\cite{monti2022stt}}
\newcommand{\MOENext}{M\nsp O\nsp E-Next\,\cite{sun2022human}}
\newcommand{\MOETrajectronpp}{M\nsp O\nsp E-\nsp Traj++\,\cite{sun2022human}\!}

\definecolor{cvprblue}{rgb}{0.21,0.49,0.74}
\usepackage[pagebackref,breaklinks,colorlinks,citecolor=cvprblue]{hyperref}

\def\paperID{3879} 
\def\confName{CVPR}
\def\confYear{2024}

\title{SingularTrajectory: Universal Trajectory Predictor Using Diffusion Model}

\author{Inhwan Bae, Young-Jae Park and Hae-Gon Jeon\thanks{Corresponding author}\\
AI Graduate School, GIST, South Korea\\
{\tt\small \{inhwanbae, youngjae.park\}@gm.gist.ac.kr, haegonj@gist.ac.kr}
}

\begin{document}
\maketitle

\begin{abstract}
There are five types of trajectory prediction tasks: deterministic, stochastic, domain adaptation, momentary observation, and few-shot. These associated tasks are defined by various factors, such as the length of input paths, data split and pre-processing methods. Interestingly, even though they commonly take sequential coordinates of observations as input and infer future paths in the same coordinates as output, designing specialized architectures for each task is still necessary. For the other task, generality issues can lead to sub-optimal performances. In this paper, we propose SingularTrajectory, a diffusion-based universal trajectory prediction framework to reduce the performance gap across the five tasks. The core of SingularTrajectory is to unify a variety of human dynamics representations on the associated tasks. To do this, we first build a Singular space to project all types of motion patterns from each task into one embedding space. We next propose an adaptive anchor working in the Singular space. Unlike traditional fixed anchor methods that sometimes yield unacceptable paths, our adaptive anchor enables correct anchors, which are put into a wrong location, based on a traversability map. Finally, we adopt a diffusion-based predictor to further enhance the prototype paths using a cascaded denoising process. Our unified framework ensures the generality across various benchmark settings such as input modality, and trajectory lengths. Extensive experiments on five public benchmarks demonstrate that SingularTrajectory substantially outperforms existing models, highlighting its effectiveness in estimating general dynamics of human movements. Code is publicly available at \url{https://github.com/inhwanbae/SingularTrajectory}.
\end{abstract}

\begin{figure}[t]
\centering
\includegraphics[width=\linewidth,trim={0 108mm 0 0mm},clip]{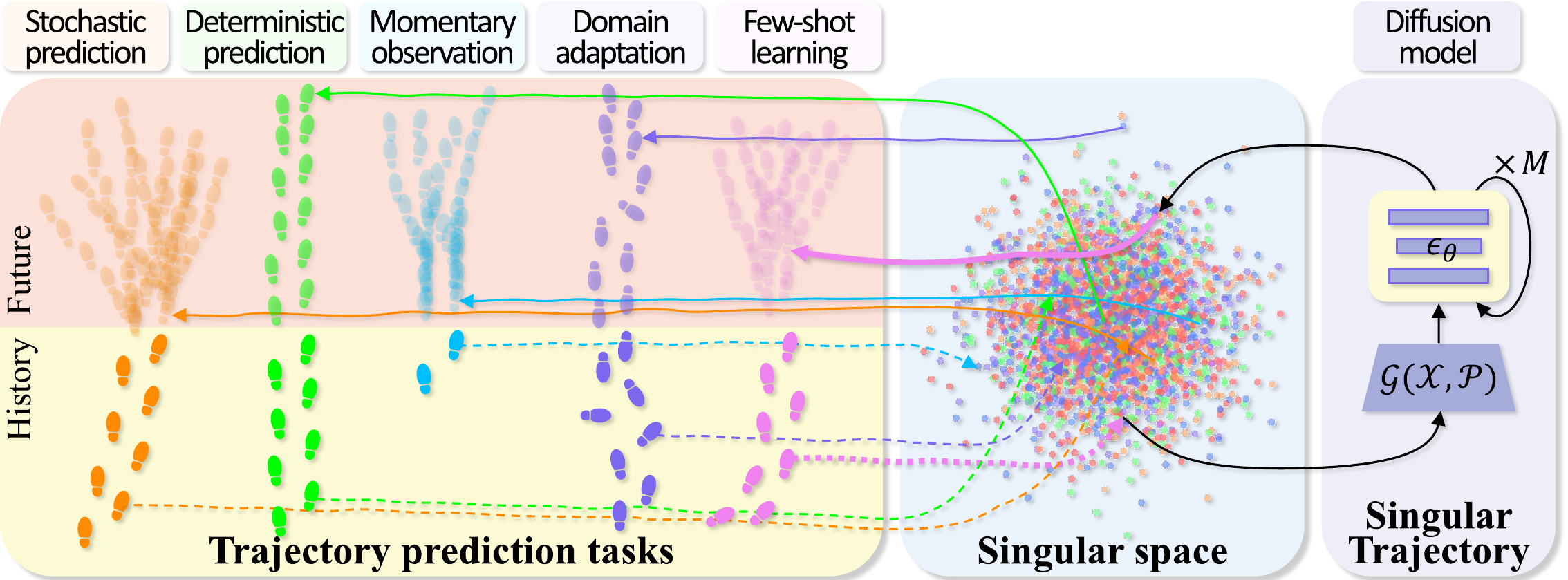}
\vspace{-6.5mm}
\caption{An overview of our SingularTrajectory framework. All relevant human trajectory prediction tasks can be represented in our Singular space, a unified feature embedding space for human dynamics. Using the embedding features, our diffusion-based model for a universal trajectory prediction makes prediction for all the tasks in this same space.}
\label{fig:teaser}
\vspace{-0.5mm}
\end{figure}

\section{Introduction}
Extensive studies of trajectory prediction methods have been conducted in the computer vision field for several decades \cite{helbing1995social,mehram2009socialforcemodel}. They have demonstrated its importance in various applications, including crowd simulation, social robot navigation, obstacle avoidance and security and surveillance systems, etc. Trajectory prediction takes sequential coordinate values of agents as input and infers their future pathways in common \cite{alahi2016social,gupta2018social,vemula2018social}. Such tasks vary depending on the application. The tasks are determined with respect to the number of input/output, data processing, and the use of geological information.

The tasks for trajectory prediction can be categorized into five groups: (1) Stochastic prediction is to predict 20 multi-modal future trajectories from one observation with 8 frames~\cite{gupta2018social}. Each future trajectory consists of 12 frames. (2) Deterministic prediction takes one observation with 8 frames, but infers only one future trajectory with 12 frames \cite{alahi2016social}. (3) The momentary observation uses only two frames to predict 20 multi-modal future paths with 12 frames \cite{sun2022human}. (4) Domain adaptation splits training data with respect to places in datasets, trains a model on one place, and then checks the transferability to other places \cite{xu2022tgnn}. (5) The few-shot task only uses partial data to build a dataset-efficient model \cite{mohamed2020social}, as illustrated in~\cref{fig:teaser}. Until now, each specialized architecture for a task type has provided performance gains. 

However, two questions arise. First, why do state-of-the-art models for one task undergo significant performance drops when applied to other trajectory prediction tasks? Second, is it feasible to design a general predictor that works across the five tasks?

As answers to these questions, we present a universal trajectory predictor to achieve generality in predictions, named SingularTrajectory. The main idea is to unify the modalities of human dynamic representations across the five tasks. To do this, we first introduce a Singular space, an embedding space consisting of representative motion patterns for each task. The motion patterns play a role in the basis function for pedestrian movements, and are extracted using Singular Value Decomposition (SVD). They are then projected onto Singular space. 

We next propose an environment-adaptive anchor working in the Singular space. Unlike traditional fixed anchor methods ~\cite{kothari2021interpretable,bae2023eigentrajectory} that sometimes fail to handle different target data distribution, our adaptive anchor is able to correct prototype paths from the adaptive anchor in Singular space if they are put into the wrong locations, based on an input traversability map. Lastly, we generate socially-acceptable future trajectories for all agents in scenes with a diffusion-based predictor which denoises residuals of perturbed prototype paths. Thanks to the cascaded denoising process of diffusion models, we refine the prototype paths. Here, historical pathways, agent interactions, and environmental information are provided as conditions to guide them in the Markov chain of the denoising diffusion processes.

Experimental results demonstrate that our SingularTrajectory can successfully represent pedestrian motion dynamics, and significantly improve prediction accuracy for the five tasks on challenging public benchmark datasets.

\section{Related Works}
We review previous studies on trajectory prediction that have attempted to address various benchmark scenarios.

\subsection{Pedestrian Trajectory Prediction}
Ways of predicting pedestrian future trajectories have been studied for a long time in the computer vision field. Pioneering works~\cite{helbing1995social,pellegrini2009you,mehram2009socialforcemodel,yamaguchi2011you} model an invisible social norm based on motion dynamics as an energy minimization problem. Introducing recurrent neural networks~\cite{alahi2016social,gupta2018social,salzmann2020trajectron++,yu2020spatio} achieves significant improvements by providing highly sequential representations of high-level shape of paths. These methods determine the most probable path, called deterministic trajectory prediction. The following works model mutual influences among agents using attention mechanisms \cite{vemula2018social,ivanovic2019trajectron,fernando2018soft,salzmann2020trajectron++}, graph convolutional networks~\cite{kipf2016semi,mohamed2020social,sun2020rsbg,bae2021dmrgcn,liu2021snce,liu2021causal,mohamed2022socialimplicit}, graph attention networks \cite{velivckovic2018graph,huang2019stgat,kosaraju2019social,liang2020garden,liang2020simaug,Shi2021sgcn,bae2022npsn}, and\,transformers\,\cite{yu2020spatio,yuan2021agent,gu2022mid,monti2022stt,bae2022gpgraph,wen2022socialode,wong2022v2net,shi2023trajectory}. Additional visual information allows us to leverage environmental constraints from traversability maps \cite{varshneya2017,xue2018sslstm,manh2018scene,sadeghian2019sophie,liang2019peeking,kosaraju2019social,sun2020reciprocal,dendorfer2020goalgan,dendorfer2021mggan,zhao2019matf,tao2020dynamic,sun2020rsbg,marchetti2020mantra,marchetti2020multiple,deo2020trajectory,shafiee2021Introvert,mangalam2021ynet,yue2022nsf,wang2023fend}. Depending on the constraints, predictors take either recurrent \cite{alahi2016social,gupta2018social,bisagno2018group,pfeiffer2018,zhang2019srlstm,xu2020cflstm,salzmann2020trajectron++,ma2020autotrajectory,zhao2021experttraj,chen2021disdis,lee2022musevae,gu2022mid,marchetti2022smemo,navarro2022social,chen2023unsupervised,maeda2023fast} or simultaneous approaches \cite{mohamed2020social,bae2021dmrgcn,Shi2021sgcn,li2021stcnet,shi2022social,xu2022remember,bae2023graphtern} to extrapolate the future pathways.

Meanwhile, with the success of generative models, the importance of multimodality has begun to emerge, called stochastic trajectory prediction. Stochastic prediction enables us to consider all of an agent's possible future pathways. For example, an agent at a crossroads may either walk straight or turn left/right. Here, the stochastic prediction infers all potential future modes. This approach has become mainstream in this field. Starting from Social-GAN~\cite{gupta2018social}, a bivariate Gaussian distribution~\cite{alahi2016social,bae2021dmrgcn,mohamed2020social,shi2020multimodal,yu2020spatio,li2020Evolvegraph,shi2021socialdpf,yao2021bitrap,Shi2021sgcn,xu2022tgnn}, generative adversarial network~\cite{gupta2018social,sadeghian2019sophie,kosaraju2019social,zhao2019matf,sun2020reciprocal,li2019idl,liang2021tpnms,dendorfer2021mggan,huang2019stgat,sun2023stimulus}, and conditional variational autoEncoder~\cite{lee2017desire,li2019conditional,ivanovic2019trajectron,bhattacharyya2020conditional,salzmann2020trajectron++,mangalam2020pecnet,chen2021disdis,sun2021pccsnet,lee2022musevae,wang2022stepwise,xu2022groupnet,xu2022socialvae} have been adopted for stochastic trajectory prediction. Anchor-conditioned methods can explicitly represent different modalities by prototyping possible paths~\cite{kothari2021interpretable,bae2023eigentrajectory}. Most recently, diffusion-based models have revealed their tremendous representation capacities in numerous tasks~\cite{sohl2015dpm,song2019eg,ho2020ddpm,song2020ddim,dhariwal2021diffusion,nichol2021iddpm}, proving its potential for stochastic trajectory prediction~\cite{gu2022mid,mao2023leapfrog,rempe2023trace,jiang2023motiondiffuser}. In this study, we take full advantage of both the anchor-conditioned approach and the diffusion-based model to achieve the explainability and generalizability of the trajectory prediction tasks.

\subsection{Various Trajectory Prediction Tasks}
Beyond the standard benchmark protocol of stochastic prediction, there are three other variants of this task: momentary observation, domain adaptation, and few-shot learning. Works in~\cite{sun2022human,monti2022stt,das2023distilling} only take two frames as input for the momentary trajectory prediction. Multi-task learning, self-supervised learning, and knowledge distillation techniques have been used to extract rich features from the limited data. Another works~\cite{xu2022tgnn,ivanovic2023expanding,zhi2023adaptive} focuses on domain adaptation across trajectory domains, captured from different surveillance views. A transferable graph neural network is introduced to adaptively learn domain-invariant knowledge. The others~\cite{mohamed2020social,zhao2021experttraj} adopt few-shot learning for better training efficiency.

Although these works take and infer the sequential coordinates of agent trajectories in common, there is no unified model for all the tasks. Despite the tremendous efforts to design a specialized architecture for one task, they cannot be applied to the other tasks without suffering significant performance drops. In the next section, we will describe how to design a unified architecture, which consistently produces promising results on the five associated tasks.

\section{Methodology}\label{sec:method}
We describe how to learn a general representation of human motions. We first define a general trajectory prediction problem in~\cref{sec:method_problem_definition} and provide preliminaries of explicit formulations on the SVD and diffusion process in~\cref{sec:method_preliminaries}. We then introduce a motion feature extraction to build our Singular space, and a projection of any trajectory from each task onto it in~\cref{sec:method_unifying}. Next, we propose an environment-adaptive anchor using motion vectors and input image in~\cref{sec:method_anchor}. Finally, we present the SingularTrajectory predictor based on the diffusion model in~\cref{sec:method_diffusion}.

\subsection{Problem Definition}\label{sec:method_problem_definition}
Trajectory prediction aims to predict the future paths of agents based on their historical path and surrounding contexts. Suppose that at each timestamp $t$, there are $N$ pedestrians in a scene with the 2D spatial coordinate position $\{\bm{p}_t^n\!\in\!\mathbb{R}^2|n\!\in\![1, ..., N]\}$. A pedestrian historical trajectory $\bm{X}_n$ over $T_{\textit{hist}}$ timesteps can be represented as the cumulative coordinates $\bm{X}_n\!=\!\{\bm{p}_t^n|t\!\in\![1, ..., T_{\textit{hist}}]\}$. Similarly, future trajectories $\bm{Y}_n$ for the time duration $T_{\textit{fut}}$ to be predicted can be written as $\bm{Y}_n\!=\!\{\bm{p}_t^n|t\!\in\![T_{\textit{hist}}\!+\!1, ..., T_{\textit{hist}}\!+\!T_{\textit{fut}}]\}$. The prediction system takes both the historical trajectories for all $N$ people $\bm{X}=\{\bm{X}_n|n\!\in\![1, ..., N]\}$ and the scene image map $\bm{I}$ for environmental information as input. The deterministic prediction system infers one sequence of the most reliable future trajectory $\widehat{\bm{Y}}=\{\widehat{\bm{Y}}_n|n\!\in\![1, ..., N]\}$. For the stochastic prediction, because of the indeterminacy of the future movements, $S$ multiple pathways for all the $N$ pedestrians $\widehat{\bm{Y}}=\{\widehat{\bm{Y}}_n^s|n\!\in\![1, ..., N], s\!\in\![1, ..., S]\}$ are generated so that at least one sample is close to the ground-truth trajectory.

\subsection{Preliminaries}\label{sec:method_preliminaries}
\vspace{0mm}\noindent\textbf{Singular Value Decomposition.}\quad
Singular Value Decomposition (SVD) decomposes a matrix into three resultant matrices. Given a matrix $\bm{A}$, its SVD is represented as:
\begin{equation}
    \bm{A} = \bm{U} \bm{\mathit{\Sigma}}\, \bm{V}^\top,
    \label{eq:svd}
\end{equation}
where $\bm{U}$ is an orthogonal left singular vector matrix, whose columns are eigenvectors of $\bm{A}\!\times\!\bm{A}^\top$. $\bm{\mathit{\Sigma}}$ is a diagonal matrix with the singular values of $\bm{A}$, consisting of $K$ non-negative values in descending order. $\bm{V}$ is a right singular vector matrix, which is also orthogonal and its columns are the eigenvectors of $\bm{A}^\top\!\times\!\bm{A}$. 

To remove the redundant part of the raw data, the truncation technique is often applied to the results after the decomposition. The idea behind the truncated SVD is to approximate the original matrix $\bm{A}$ with the lower rank. With the $K$ to determine the number of singular values to retain, $\bm{\mathit{\Sigma}}$ can be simplified to $\bm{\mathit{\Sigma}}_K$ which contains only the $K$ largest singular values. Similarly, $\bm{U}$ and $\bm{V}^\top$ are reduced to $\bm{U}_K$ and $\bm{V}^\top_K$ by keeping the first $K$ columns and rows, respectively. This process eliminates the smallest singular values, which are not needed to express the original data and often correspond to noise or redundant information. This is useful for practical scenarios dealing with large and potentially sparse matrices because we can reconstruct a close approximation of the original data with significantly less storage space. 

\vspace{0.5mm}\noindent\textbf{Diffusion models.}\quad
The diffusion model operates by transforming a noisy distribution, represented by the noise vector $\bm{\mathrm{y}}_M$, into the desired data $\bm{\mathrm{y}}_0$ through a series of $M$ diffusion steps. These steps involve intermediate latent variables $\{\bm{\mathrm{y}}_m|m\in[1,\ldots,M]\}$, and encompass both the diffusion and denoising processes. The diffusion process adds a small amount of noise to data in order to obtain the standard normal distribution $q(\bm{\mathrm{y}}_M)$ from the distribution $q(\bm{\mathrm{y}}_0)$ using the Markov chain as:
\begin{equation}\label{eq:diffusion_forward}
\begin{gathered}
q(\bm{\mathrm{y}}_{1:M} | \bm{\mathrm{y}}_0) := \prod_{m=1}^{M} q(\bm{\mathrm{y}}_m | \bm{\mathrm{y}}_{m-1}) \vspace{-2pt}\\
\quad q(\bm{\mathrm{y}}_m | \bm{\mathrm{y}}_{m-1}) := \mathcal{N}\big(\bm{\mathrm{y}}_m; \sqrt{1-\beta_m} \bm{\mathrm{y}}_{m-1}, \beta_m \bm{\mathrm{I}}\big) 
\end{gathered}
\end{equation}
where $\beta_t$ is a small positive constant and a variance schedule for adding noise. The denoising process uses the $\bm{\mathrm{y}}_m$ to recover $\bm{\mathrm{y}}_0$ with a learnable network as follows:
\begin{equation}\label{eq:diffusion_reverse}
\begin{gathered}
p_\theta(\bm{\mathrm{y}}_{0:M}) := p(\bm{\mathrm{y}}_M) \prod_{m=1}^{M} p_\theta(\bm{\mathrm{y}}_{m-1} | \bm{\mathrm{y}}_m), \vspace{-2pt}\\
p_\theta(\bm{\mathrm{y}}_{m-1} | \bm{\mathrm{y}}_m) := \mathcal{N}(\bm{\mathrm{y}}_{m-1}; \bm{\epsilon}_\theta(\bm{\mathrm{y}}_m,m),\beta_m\bm{\mathrm{I}}).
\end{gathered}
\end{equation}
where $\bm{\mathrm{y}}_M \sim \mathcal{N}(\bm{0},\bm{\mathrm{I}})$ is an initial noise sampled from the Gaussian distribution $p(\bm{\mathrm{y}}_M)$, and $\theta$ denotes the learnable parameter of the diffusion model. $\bm{\epsilon}_\theta$ is a learnable denoising model of a clean data\,$\bm{\mathrm{y}}_{0}$, and a corrupted data $\bm{\mathrm{y}}_{m}$\,at a step $m$. The objective is to train the neural network so that the denoising process predicts the true data-generating distribution well. This is often done by maximizing the evidence lower bound, ensuring that the samples generated by the diffusion model are indistinguishable from the real data.

\subsection{Unifying the Motion Space}\label{sec:method_unifying}
The trajectory prediction model uses a learnable network to capture the relationship in consideration of the input coordinates of pedestrians, input images and output coordinates for each pedestrian. Since expectations for input and output trajectories are different for each associated task (\eg, length and multimodality), they should be viewed as different data spaces even if they use the same coordinate systems. We introduce a method to merge raw data in each space into a Singular space for human motion dynamics.

\vspace{0.5mm}\noindent\textbf{Singular space construction.}\quad
To discover motion dynamics from the raw data, we first extract primitive motion features. Inspired by the successful low-rank approximation of raw trajectory data using eigenvectors in~\cite{bae2023eigentrajectory}, we also employ a similar strategy using singular vectors from the truncated SVD to extract principal motion components from the entire training dataset.

First, we cut-off paths of all pedestrians in the dataset into $T_{\textit{win}}$ lengths through a sliding window to create a total of $L$ gist of trajectory set $\bm{A}\in\mathbb{R}^{L\times(2\times T_{\textit{win}})}$. Here, $\bm{A}$ is a temporary matrix to extract a motion vector from a set of trajectory in the initialization phase. Next, we decompose $\bm{A}$ to obtain truncated matrices $\bm{U}_K\in\mathbb{R}^{L\times K}$, $\bm{\mathit{\Sigma}}_K\in\mathbb{R}^{K\times K}$ and $\bm{V}^\top_K\in\mathbb{R}^{K\times(2\times T_{\textit{win}})}$. Here, $\bm{V}^\top_K$ is a set of $K$ spatio-temporal motion vectors $\mathrm{v}_k\in\mathbb{R}^{2\times T_{\textit{win}}}$ representing the most dominant motion dynamics of pedestrians. Since $\mathrm{v}_k$ are orthogonal to each other, we define a Singular space coordinate system using the $K$ singular vectors as basis vectors for each axis. In Singular space, trajectories $\bm{A}$ can be a coordinate of $\mathcal{A}=\bm{U}_K \cdot \bm{\mathit{\Sigma}}_K \in\mathbb{R}^{L\times K}$, indicating that each motion vector has an influence on reconstructing the $L$ trajectories. In other words, we can project the $\bm{A}$ into the coordinate in Singular space $\mathcal{A}$ as follows: 
\begin{equation}\label{eq:projection}
\mathcal{A}=\bm{A}\times (\bm{V}^\top_K)^{-1}=\bm{A}\times\bm{V}_K,
\end{equation}
where $(\bm{V}^\top_K)^{-1}$ can be simplified into $\bm{V}_K$ with the property of an orthogonal matrix\footnote{For better understanding, we display the coordinate variable in the Singular space using the calligraphic font}.

In the Singular space, we can now concentrate on the motion flow by using the coefficients of motion patterns, instead of considering every position over time. Note that the results from the SVD depend highly on various factors such as the window size for inputs/outputs, data processing, and the differences in the raw data caused by geological variations. Due to the issue, works using SVD only convert the output space~\cite{jin2022eigenlanes,park2022eigencontours,turk1991eigenfaces} or conduct separate decompositions for each of input and output data spaces~\cite{bae2023eigentrajectory}. However, in this case, they cause inconsistent representations of motion dynamics, even for the same pedestrian, which leads to a lack of generality.
In the next step, we introduce a new method to address this issue.

\begin{figure}[t]
\centering
\includegraphics[width=\linewidth,trim={0 87mm 0 0},clip]{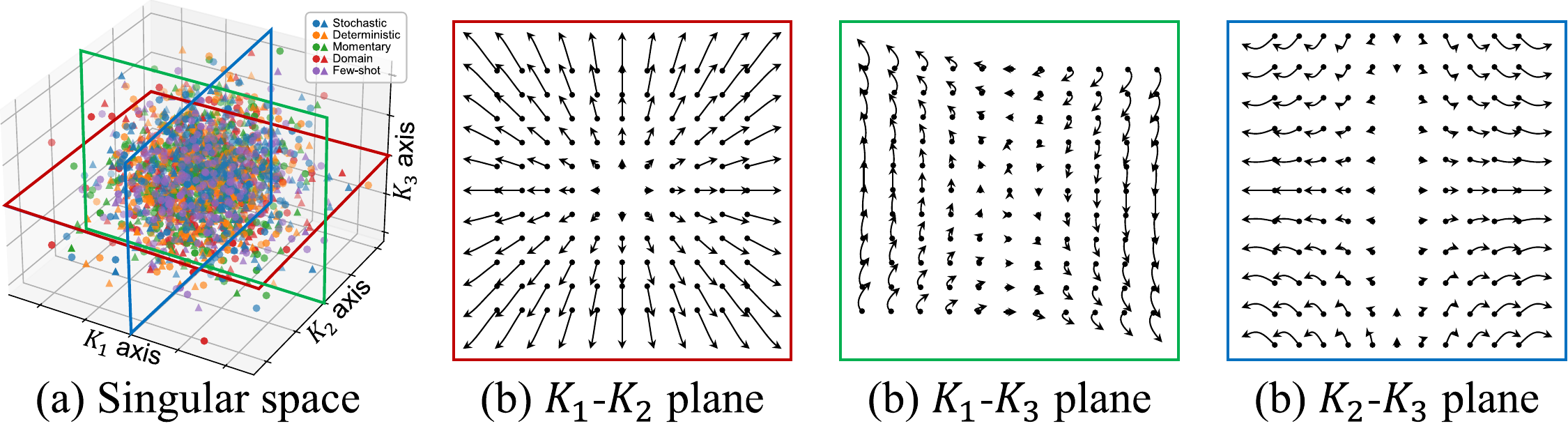}
\vspace{-6.5mm}
\caption{Visualization of the trajectories in Singular space. (a) The circle and triangle markers indicate the history and future trajectory coordinates, respectively. Each color also refers to each associated task. (b-d) Each marker in the Singular space corresponds to each trajectory in raw data. And, the slicing planes, representing a set of arrows, mean human dynamics of straight forward motions and turning motions.}
\label{fig:method_singular_space}
\end{figure}

\vspace{0.5mm}\noindent\textbf{Projection of any trajectory into Singular space.}\quad
We aim to express the input/output trajectories $\bm{X}$ and $\bm{Y}$ of the five associated tasks in Singular space all at once. Our core idea stems from the notion that all pedestrians share the same human motion dynamics and thus will likely show a similar pattern. Starting with the projection function \cref{eq:projection}, which is a projection matrix for the fixed trajectory length $T_{\textit{win}}$, we extend it to any length $T_{\textit{hist}}$ and $T_{\textit{fut}}$, which varies from task to task. To handle the motion patterns regardless of their lengths, we interpolate $\mathrm{v}_k\in\mathbb{R}^{2\times T_{\textit{win}}}$ to $\mathrm{v}_{\mathrm{x},k}\in\mathbb{R}^{2\times T_{\textit{hist}}}$. Since the motion pattern can be seen as a 2-dimensional curve, we use Cardinal B-splines to make a transformation matrix $\bm{C}_{T_{\textit{hist}}}\in\mathbb{R}^{(2\times T_{\textit{hist}})\times (2\times T_{\textit{win}})}$ using the Irwin-Hall distribution. The constant value $\bm{C}$ depends only on the length, so $\mathrm{v}_{\mathrm{x},k}$ can be approximated. The trajectory $\bm{X}\in\mathbb{R}^{N\times(2\times T_{\textit{hist}})}$ is projected to the coordinate $\mathcal{X}\in\mathbb{R}^{N\times(2\times T_{\textit{hist}})}$ in Singular space as follows:
\begin{equation}
\mathrm{v}_{\mathrm{x},k} = \bm{C}_{T_{\textit{hist}}}\times\mathrm{v}_k.
\vspace{-5mm}
\end{equation}
\begin{equation}\label{eq:singular_projection}
\mathcal{X} = \bm{X}\times\bm{C}_{T_{\textit{hist}}}\times\bm{V}_K.
\end{equation}
In the same way as \cref{eq:singular_projection}, the trajectory $\bm{Y}\in\mathbb{R}^{N\times(2\times T_{\textit{fut}})}$ also can be projected to $\mathcal{Y}\in\mathbb{R}^{N\times K}$ using $\bm{V}_K$ and $\bm{C}_{T_{\textit{fut}}}$.

Through this process, Singular space can represent a variety of trajectories even with different lengths, including motion vectors for the input/output as well as the task-specific data, as visualized in~\cref{fig:method_singular_space}. Establishing a common ground for the trajectory representation is crucial for our model's adaptability and robust performances across different trajectory prediction benchmarks. Moreover, the strategy of focusing on the overall motion flow, rather than frame-by-frame coordinates, further improves the model's capacity to understand and predict socially compliant trajectories in diverse real-world scenarios.

\begin{figure}[t]
\centering
\includegraphics[width=\linewidth,trim={0 78mm 0 0},clip]{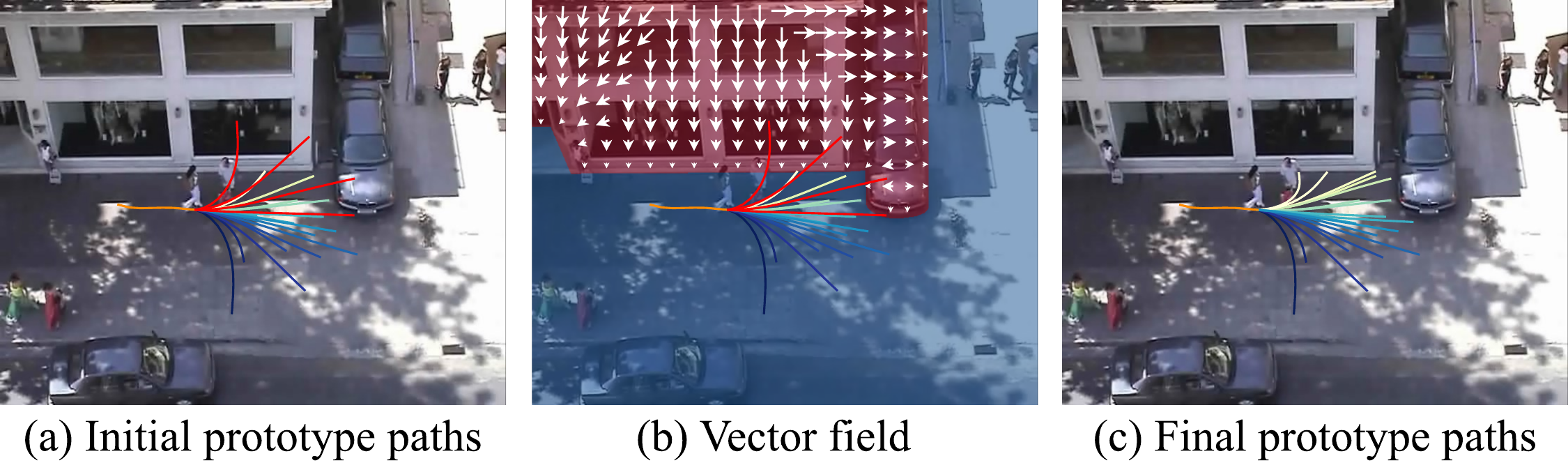}
\vspace{-6.5mm}
\caption{An example of the adaptive anchor generation. (a) The initial prototype anchor $\mathcal{P}$ is placed on the last observation coordinate of a person. In this instance, four prototype paths (highlighted in red) are incorrectly placed at the non-traversable locations. (b) Vector field $\vec{F}_{I}(x,y)$ is computed to guide toward in the nearest traversable areas. (c) The initial prototype paths are then tailored to the environment using the vector field.}
\label{fig:method_adaptive_anchor}
\end{figure}

\subsection{Adaptive Anchor}\label{sec:method_anchor}
Beyond the trajectory data integration, we also introduce how to incorporate environmental contexts into trajectory predictors. To do this, we propose an adaptive anchor which is expressed as a set of motion vectors from input images and consists of prototype paths in Singular space. 

\vspace{0.5mm}\noindent\textbf{Prototype anchor formation.}\quad
Previous anchor-based human trajectory prediction approaches~\cite{kothari2021interpretable,bae2023eigentrajectory} use a fixed anchor for any pedestrian, and the anchor is refined for output trajectories. Although these methods can explicitly model the multimodality, they sometimes fail to handle the case where the prototype path is put into wrong locations, or blocked by static obstacles. This is because the wrong prototypes are treated as a hard constraint. To avoid this problem, we use the input image as a traversability map to correct the wrong prototype paths.

We start with the converted future trajectory $\mathcal{Y}$ in the whole dataset $\bm{Y}$ using \cref{sec:method_unifying}. Following~\cite{park2022eigencontours,bae2023eigentrajectory}, the coordinates $\mathcal{Y}$ are then clustered into $S$ centroids, which can be viewed as groups of components representing different multimodal futures. In addition, each centroid in Singular space is used as an initial prototype path $\mathcal{P}_s$ for constructing an anchor $\mathcal{P}\in\mathbb{R}^{S\times K}$. But, these initial prototype paths are still fixed, so they still have a limitation which is unlikely to consider environmental information. 

\vspace{0.5mm}\noindent\textbf{Adaptive anchor generation.}\quad
To make use of environmental information, we introduce a module to deform the anchor. Using an off-the-shelf semantic segmentation model from~\cite{mangalam2021ynet}, we can obtain binary traversable maps $\bm{I}_{\textit{map}}$ from an input image. We then derive a vector field $\vec{F}_{I}(x,y)$ to fix the wrongly located prototype path by directing it to a nearest traversable regions. If the initial prototype path $\mathcal{P}_s$ is in the wrong place, we deform it by adding the vector fields into the initial anchor until they reach equilibrium states to obtain the final prototype paths $\mathcal{P}'_s$ as follows:
\begin{equation}
    \mathcal{P}'_s = \mathcal{P}_s + \vec{F}_{I}(\mathcal{P}_s\bm{V}^\top_K\bm{C}_{T_{\textit{fut}}}^{-1}).
\end{equation}
Through this process, the prototype path is re-located toward the nearest walkway, as demonstrated in \cref{fig:method_adaptive_anchor}. In other words, the anchor plays a role in the environment-adaptive prototype paths, unlike the existing fixed anchor methods.

Because the prototype paths are in Singular space, the scene image can be implicitly represented as the adaptive anchor. In addition, the scene image can be approximated with the set of motions. By projecting the scene images into the Singular space, our model understands the surrounding environment. This holistic approach, integrating both coordinate and environmental cues, sets the stage for more realistic and reliable trajectory predictions.

\subsection{Diffusion-Based SingularTrajectory Model}\label{sec:method_diffusion}
As the final step, we develop a framework called SingularTrajectory, a unified model that works well across the five tasks. Leveraging the Singular space and the adaptive anchor, our diffusion-based SingularTrajectory model can precisely forecast potential future paths.

\vspace{0.5mm}\noindent\textbf{Denoising a perturbed trajectory anchor.}\quad
Unlike previous diffusion-based predictors, which directly forecast the future path from Gaussian noise~\cite{gu2022mid,jiang2023motiondiffuser}, we devise a stepwise refinement from the adaptive anchor for a realistic trajectory. Here, historical pathways $\mathcal{X}$, environmental information $\mathcal{P}$, and agent interactions $\mathcal{G}(\mathcal{X},\mathcal{P})$ are encoded and used as conditions to guide the denoising processes. To better capture agents interactions $\mathcal{G}$ in the denoising process, we adopt the transformer model. Similar to other transformer-based trajectory predictors~\cite{gu2022mid,mao2023leapfrog}, our SingularTrajectory encodes spatio-temporal information to account for agent-agent and agent-environment interactions using $\mathcal{X}$ and $\mathcal{P}$, respectively. These conditions are then concatenated into one feature vector representation and fed into the diffusion model $\bm{\epsilon}_\theta(\bm{\mathrm{y}}_m, m, \mathcal{X}, \mathcal{P}, \mathcal{G})$ to learn, by contrasting the motion patterns from previous diffusion steps to distinguish the added noise, and to close the reality gap by generating socially-acceptable future paths $\widehat{\bm{Y}}$.

This refinement process works in a cascading manner by $\{\mathcal{P}, ..., \widehat{\mathcal{Y}}\}_{m=1}^M$ as follows:
\begin{equation}
    \widehat{\mathcal{Y}} = \mathcal{P}+p(\bm{\mathrm{y}}_M) \smash{\prod_{m=1}^{M}} p_\theta(\bm{\mathrm{y}}_{m-1} | \bm{\mathrm{y}}_m).
\end{equation}
By predicting only the residuals $\bm{\mathrm{y}}_m\in\mathbb{R}^{S\times K}$ to adjust the anchors, the problem is simplified, in that the model is able to use a prior knowledge on the initial state. With this process, we thus ensure more precise and reliable trajectory generations, achieving generality in various environments and applications.

\vspace{0.5mm}\noindent\textbf{Implementation details.}\quad
To construct Singular space, we empirically set $K$ to 4. We set $T_{\textit{win}}\!=\!T_{\textit{fut}}\!=\!12$ in order to prevent information loss due to the approximation of the motion vector during prediction. For an anchor diffusion model, we devise a one-layer transformer for encoding motion and context information, where the dimension is set to 256 with 4-head attention. We set $M=10$ and schedule the diffusion timesteps following DDIM~\cite{song2020ddim}. To train the SingularTrajectory in an end-to-end manner, we use a mean square error (MSE) as a loss function between the output and a random Gaussian noise for the current iteration. The training is performed with AdamW optimizer~\cite{loshchilov2018decoupled}, with a learning rate of 0.001 and batch size of 512 for 256 epochs. All the experiments are conducted on a single NVIDIA A6000 GPU, which usually takes about an hour to train each scene.

\begin{table*}[t]
\Large
\centering
\resizebox{\linewidth}{!}{%
\begin{tabular}{ccc@{\,\,}c@{\,\,}c@{\,\,}cccc@{\,\,}c@{\,\,}c@{\,\,}cccc@{\,\,}c@{\,\,}c@{\,\,}cccc@{\,\,}c@{\,\,}c@{\,\,}cccc@{\,\,}c@{\,\,}c@{\,\,}cccc}
\toprule
\multirow{2}{*}{\tworow{Domain Adaptation}{(ADE)}\vspace{-5pt}} & & \multicolumn{5}{c}{A2*} & & \multicolumn{5}{c}{B2*} & & \multicolumn{5}{c}{C2*} & & \multicolumn{5}{c}{D2*} & & \multicolumn{5}{c}{E2*} & & *2* \\   \cmidrule(lr){3-7} \cmidrule(lr){9-13} \cmidrule(lr){15-19} \cmidrule(lr){21-25} \cmidrule(lr){27-31} \cmidrule(lr){33-33}
                     & & A2B  & A2C  & A2D  & A2E  & AVG  & & B2A  & B2C  & B2D  & B2E  & AVG  & & C2A  & C2B  & C2D  & C2E  & AVG  & & D2A  & D2B  & D2C  & D2E  & AVG  & & E2A  & E2B  & E2C  & E2D  & AVG  & & AVG  \\ \midrule
\STGCNN              & & 1.83 & 1.58 & 1.30 & 1.31 & 1.51 & & 3.02 & 1.38 & 2.63 & 1.58 & 2.15 & & 1.16 & 0.70 & 0.82 & 0.54 & 0.81 & & 1.04 & 1.05 & 0.73 & 0.47 & 0.82 & & 0.98 & 1.09 & 0.74 & 0.50 & 0.83 & & 1.22 \\
\PECNet              & & 1.97 & 1.68 & 1.24 & 1.35 & 1.56 & & 3.11 & 1.35 & 2.69 & 1.62 & 2.19 & & 1.39 & 0.82 & 0.93 & 0.57 & 0.93 & & 1.10 & 1.17 & 0.92 & 0.52 & 0.93 & & 1.01 & 1.25 & 0.83 & 0.61 & 0.93 & & 1.31 \\
\AgentFormer         & & 1.73 & 1.64 & 1.87 & 1.57 & 1.70 & & 1.49 & 1.76 & 2.13 & 1.73 & 1.78 & & 1.73 & 0.19 & 2.01 & 1.79 & 1.43 & & 1.71 & 2.05 & 1.92 & 1.32 & 1.75 & & 1.96 & 1.91 & 1.89 & 1.91 & 1.92 & & 1.72 \\
\MID                 & & 1.26 & 1.43 & 1.65 & 1.43 & 1.44 & & 1.23 & 0.87 & 0.98 & 0.77 & 0.96 & & 1.61 & 0.63 & 0.81 & 0.58 & 0.91 & & 1.42 & 0.75 & 0.81 & 0.74 & 0.93 & & 1.47 & 0.68 & 0.88 & 0.91 & 0.98 & & 1.04 \\
\ETFullCompact       & & 0.39 & 0.71 & 0.73 & \tul{0.45} & 0.57 & & 0.95 & 0.68 & 0.51 & 0.42 & 0.64 & & \tul{0.94} & \tul{0.32} & 0.49 & 0.37 & \tul{0.53} & & 1.10 & 1.03 & 1.36 & 0.43 & 0.98 & & 0.92 & 0.51 & 0.62 & 0.52 & 0.64 & & 0.67 \\ \cmidrule(lr){1-33}
\TGNN                & & 1.13 & 1.25 & 0.94 & 1.03 & 1.09 & & 2.54 & 1.08 & 2.25 & 1.41 & 1.82 & & 0.97 & 0.54 & 0.61 & \tbf{0.23} & 0.59 & & \tul{0.88} & 0.78 & 0.59 & \tbf{0.32} & \tul{0.64} & & \tul{0.87} & 0.72 & 0.65 & \tbf{0.34} & 0.65 & & 0.96 \\
\Kzero               & & 0.45 & 0.78 & 0.68 & 0.59 & 0.63 & & \tul{0.87} & 0.65 & \tul{0.49} & \tul{0.36} & \tul{0.59} & & 1.08 & 0.46 & \tbf{0.44} & \tul{0.36} & 0.59 & & 1.13 & 0.56 & 0.59 & 0.53 & 0.70 & & 1.13 & 0.39 & 0.59 & 0.48 & 0.65 & & 0.63 \\
\Adaptive            & & \tul{0.36} & \tul{0.64} & \tul{0.53} & \tul{0.45} & \tul{0.50} & & 0.90 & \tul{0.62} & 0.50 & \tul{0.36} & 0.60 & & 1.11 & 0.46 & \tul{0.48} & 0.42 & 0.62 & & 1.11 & \tul{0.52} & \tbf{0.54} & 0.47 & 0.66 & & 1.05 & \tul{0.34} & \tul{0.56} & \tul{0.44} & \tul{0.60} & & \tul{0.59} \\ \cmidrule(lr){1-33}
\tbf{\STFullCompact} & & \tbf{0.29} & \tbf{0.59} & \tbf{0.51} & \tbf{0.42} & \tbf{0.45} & & \tbf{0.66} & \tbf{0.55} & \tbf{0.45} & \tbf{0.35} & \tbf{0.50} & & \tbf{0.79} & \tbf{0.30} & \tul{0.48} & \tul{0.36} & \tbf{0.48} & & \tbf{0.73} & \tbf{0.29} & \tul{0.55} & \tul{0.36} & \tbf{0.49} & & \tbf{0.64} & \tbf{0.23} & \tbf{0.55} & 0.45 & \tbf{0.47} & & \tbf{0.48} \\
\bottomrule
\toprule
\multirow{2}{*}{\tworow{Domain Adaptation}{(FDE)}\vspace{-5pt}} & & \multicolumn{5}{c}{A2*} & & \multicolumn{5}{c}{B2*} & & \multicolumn{5}{c}{C2*} & & \multicolumn{5}{c}{D2*} & & \multicolumn{5}{c}{E2*} & & *2* \\   \cmidrule(lr){3-7} \cmidrule(lr){9-13} \cmidrule(lr){15-19} \cmidrule(lr){21-25} \cmidrule(lr){27-31} \cmidrule(lr){33-33}
                     & & A2B  & A2C  & A2D  & A2E  & AVG  & & B2A  & B2C  & B2D  & B2E  & AVG  & & C2A  & C2B  & C2D  & C2E  & AVG  & & D2A  & D2B  & D2C  & D2E  & AVG  & & E2A  & E2B  & E2C  & E2D  & AVG  & & AVG  \\ \midrule
\STGCNN              & & 3.24 & 2.86 & 2.53 & 2.43 & 2.77 & & 5.16 & 2.51 & 4.86 & 2.88 & 3.85 & & 2.30 & 1.34 & 1.74 & 1.10 & 1.62 & & 2.21 & 1.99 & 1.41 & 0.88 & 1.62 & & 2.10 & 2.05 & 1.47 & 1.01 & 1.66 & & 2.30 \\
\PECNet              & & 3.33 & 2.83 & 2.53 & 2.45 & 2.79 & & 5.23 & 2.48 & 4.90 & 2.86 & 3.87 & & 2.22 & 1.32 & 1.68 & 1.12 & 1.59 & & 2.20 & 2.05 & 1.52 & 0.88 & 1.66 & & 2.10 & 1.84 & 1.45 & 0.98 & 1.59 & & 2.29 \\
\AgentFormer         & & 3.60 & 3.54 & 4.52 & 3.95 & 3.90 & & 3.76 & 3.67 & 5.17 & 3.42 & 4.01 & & 3.77 & 3.49 & 4.12 & 3.76 & 3.79 & & 3.88 & 3.75 & 4.27 & 3.61 & 3.88 & & 3.82 & 3.55 & 3.74 & 4.32 & 3.86 & & 3.89 \\
\MID                 & & 2.42 & 2.77 & 3.22 & 2.77 & 2.80 & & 2.34 & 1.66 & 1.87 & 1.44 & 1.83 & & 3.13 & 1.28 & 1.74 & 1.24 & 1.85 & & 2.73 & 1.49 & 1.67 & 1.54 & 1.86 & & 2.87 & 1.41 & 1.88 & 1.93 & 2.02 & & 2.07 \\
\ETFullCompact       & & 0.76 & 1.43 & 1.56 & 0.96 & 1.18 & & 2.10 & 1.38 & 1.07 & 0.86 & 1.35 & & 2.01 & \tul{0.64} & 1.10 & 0.82 & \tul{1.14} & & 2.42 & 1.03 & 1.36 & 0.97 & 1.44 & & 2.04 & \tul{0.51} & 1.29 & 1.11 & 1.24 & & 1.27 \\ \cmidrule(lr){1-33}
\TGNN                & & 2.18 & 2.25 & 1.78 & 1.84 & 2.01 & & 4.15 & 1.82 & 4.04 & 2.53 & 3.14 & & \tul{1.91} & 1.12 & 1.30 & 0.87 & 1.30 & & \tul{1.92} & 1.46 & 1.25 & \tbf{0.65} & \tul{1.32} & & \tul{1.86} & 1.45 & 1.28 & \tbf{0.72} & 1.33 & & 1.82 \\
\Kzero               & & 0.94 & 1.72 & 1.51 & 1.29 & 1.37 & & \tul{1.81} & 1.33 & \tul{1.02} & \tul{0.75} & \tul{1.23} & & 2.17 & 0.91 & \tbf{0.94} & \tul{0.81} & 1.21 & & 2.18 & \tul{0.98} & 1.20 & 1.03 & 1.35 & & 2.12 & 0.68 & 1.22 & 1.03 & 1.26 & & 1.28 \\
\Adaptive            & & \tul{0.69} & \tul{1.34} & \tul{1.12} & \tul{0.91} & \tul{1.02} & & 1.85 & \tul{1.27} & 1.05 & 0.76 & \tul{1.23} & & 2.26 & 0.92 & \tul{1.03} & 0.94 & 1.29 & & 2.23 & 1.02 & \tbf{1.17} & 1.00 & 1.36 & & 2.07 & 0.63 & \tul{1.19} & \tul{0.94} & \tul{1.21} & & \tul{1.22} \\ \cmidrule(lr){1-33}
\tbf{\STFullCompact} & & \tbf{0.57} & \tbf{1.19} & \tbf{1.08} & \tbf{0.81} & \tbf{0.91} & & \tbf{1.16} & \tbf{1.17} & \tbf{0.97} & \tbf{0.73} & \tbf{1.01} & & \tbf{1.36} & \tbf{0.62} & 1.11 & \tbf{0.79} & \tbf{0.97} & & \tbf{1.22} & \tbf{0.52} & \tul{1.18} & \tul{0.73} & \tbf{0.92} & & \tbf{1.07} & \tbf{0.43} & \tbf{1.15} & \tul{0.94} & \tbf{0.90} & & \tbf{0.94} \\
\bottomrule
\end{tabular}
}
\vspace{-3mm}
\caption{Comparison of our SingularTrajectory with other state-of-the-art methods in domain adaptation task. Following~\cite{xu2022tgnn,ivanovic2023expanding}, we deterministically predict one future path for all dataset pairs (ADE/FDE). \tbf{Bold}: Best, \tul{Underline}: Second best.}
\label{tab:result_domain}
\vspace{-2mm}
\end{table*}

\begin{table}[t]
\Large
\centering
\resizebox{\linewidth}{!}{%
\begin{tabular}{ccccccc}
    \toprule
    Stochastic & ETH & HOTEL & UNIV & ZARA1 & ZARA2 & AVG \\ \midrule
    \SocialGAN           & 0.87\pslp1.62 & 0.67\pslp1.37 & 0.76\pslp1.52 & 0.35\pslp0.68 & 0.42\pslp0.84 & 0.61\pslp1.21 \\
    \STGCNN              & 0.64\pslp1.11 & 0.49\pslp0.85 & 0.44\pslp0.79 & 0.34\pslp0.53 & 0.30\pslp0.48 & 0.44\pslp0.75 \\
    \STAR                & 0.57\pslp1.11 & 0.19\pslp0.37 & 0.35\pslp0.75 & 0.26\pslp0.57 & 0.25\pslp0.58 & 0.33\pslp0.68 \\
    \PECNet              & 0.61\pslp1.07 & 0.22\pslp0.39 & 0.34\pslp0.56 & 0.25\pslp0.45 & 0.19\pslp0.33 & 0.32\pslp0.56 \\
    \MID                 & 0.57\pslp0.93 & 0.21\pslp0.33 & 0.29\pslp0.55 & 0.28\pslp0.50 & 0.20\pslp0.37 & 0.31\pslp0.54 \\
    \LBEBM               & 0.60\pslp1.06 & 0.21\pslp0.38 & 0.28\pslp0.54 & 0.21\pslp0.39 & 0.15\pslp0.30 & 0.29\pslp0.53 \\
    \Trajectronpp        & 0.61\pslp1.03 & 0.20\pslp0.28 & 0.30\pslp0.55 & 0.24\pslp0.41 & 0.18\pslp0.32 & 0.31\pslp0.52 \\
    \GroupNet            & 0.46\pslp0.73 & 0.15\pslp0.25 & 0.26\pslp0.49 & 0.21\pslp0.39 & 0.17\pslp0.33 & 0.25\pslp0.44 \\
    \AgentFormer         & 0.46\pslp0.80 & 0.14\pslp0.22 & 0.25\pslp0.45 & \tul{0.18}\pslp0.30 & \tul{0.14}\pslp0.24 & \tul{0.23}\pslp0.40 \\
    \GPGraph             & 0.43\pslp0.63 & 0.18\pslp0.30 & 0.24\pslp0.42 & \tbf{0.17}\pslp0.31 & 0.15\pslp0.29 & \tul{0.23}\pslp0.39 \\
    \NPSN                & \tul{0.36}\pslp0.59 & 0.16\pslp0.25 & \tul{0.23}\pslp\tul{0.39} & \tul{0.18}\pslp0.32 & \tul{0.14}\pslp0.25 & \tbf{0.21}\pslp0.36 \\
    \EqMotion            & 0.40\pslp0.61 & \tul{0.12}\pslp\tul{0.18} & \tul{0.23}\pslp0.43 & \tul{0.18}\pslp0.32 & \tbf{0.13}\pslp\tul{0.23} & \tbf{0.21}\pslp0.35 \\
    \ETFullCompact       & \tul{0.36}\pslp\tul{0.53} & \tul{0.12}\pslp0.19 & 0.24\pslp0.43 & 0.19\pslp0.33 & \tul{0.14}\pslp0.24 & \tbf{0.21}\pslp0.34 \\
    \SocialVAE           & 0.41\pslp0.58 & 0.13\pslp0.19 & \tbf{0.21}\pslp\tbf{0.36} & \tbf{0.17}\pslp\tul{0.29} & \tbf{0.13}\pslp\tbf{0.22} & \tbf{0.21}\pslp\tul{0.33} \\
    \LED                 & 0.39\pslp0.58 & \tbf{0.11}\pslp\tbf{0.17} & 0.26\pslp0.43 & \tul{0.18}\pslp\tbf{0.26} & \tbf{0.13}\pslp\tbf{0.22} & \tbf{0.21}\pslp\tul{0.33} \\
    \cmidrule(lr){1-7}
    \tbf{\STFullCompact} & \tbf{0.35}\pslp\tbf{0.42} & 0.13\pslp0.19 & 0.25\pslp0.44 & 0.19\pslp0.32 & 0.15\pslp0.25 & \tbf{0.21}\pslp\tbf{0.32} \\ 
    \bottomrule
\end{tabular}
}
\vspace{-3mm}
\caption{Evaluation on the stochastic trajectory prediction task.}
\label{tab:result_stochastic}
\end{table}

\section{Experiments}\label{sec:experiment}
In this section, we conduct comprehensive experiments to verify the generality of our SingularTrajectory model for the trajectory prediction tasks. We first describe our experimental setup in~\cref{sec:experiment_setup}. We then provide comparison results with state-of-the-art models on public benchmark datasets in~\cref{sec:experiment_result}. Finally, we perform an extensive ablation study demonstrating the effect of each component of our method in \cref{sec:experiment_ablation}.

\subsection{Experimental Setup}\label{sec:experiment_setup}
\vspace{0mm}\noindent\textbf{Datasets.}\quad
To compare our SingularTrajectory with state-of-the-art baselines, we conduct quantitative evaluations on two common datasets, ETH~\cite{pellegrini2009you} and UCY~\cite{lerner2007crowdsbyexample}, for all the associated tasks. The ETH and UCY datasets consist of various motions of 1,536 pedestrians across five unique scenes: ETH, Hotel, Univ, Zara1 and Zara2. They are recorded in surveillance views, and the trajectories are labeled in world coordinates.

\vspace{0.5mm}\noindent\textbf{Benchmarks.}\quad
We evaluate our model for the five trajectory prediction tasks: (1) stochastic prediction, (2) deterministic prediction, (3) momentary observation, (4) domain adaptation, and (5) few-shot learning. The stochastic trajectory prediction employs a leave-one-out strategy~\cite{gupta2018social,huang2019stgat,mohamed2020social,Shi2021sgcn,bae2022npsn} for training and inference across the five ETH-UCY scenes. In the scenes, the initial 3.2 seconds (equivalent to $T_{\textit{hist}}\!=\!8$ frames) of all paths are used as input, and the subsequent 4.8 seconds (corresponding to $T_{\textit{fut}}\!=\!12$ frames) are predicted. The model generates $S\!=\!20$ paths, and one with the lowest error is chosen as the final prediction. In the deterministic prediction, the trajectory prediction methods generate only $S\!=\!1$ path. For the momentary observation, the observation length is $T_{\textit{hist}}\!=\!2$ whereas still $T_{\textit{fut}}\!=\!12$. For the domain adaptation, models are trained on each of the five scenes individually, and then evaluated on the other scenes, instead of using the leave-one-out strategy. In the few-shot learning, they are trained using only 10\% of the training set, and then assessed on the entire test set. To measure the prediction performances, two common metrics are used for all the tasks, Average Displacement Error (ADE) and Final Displacement Error (FDE), measuring the average and destination error between prediction and its ground-truth path, respectively.

\begin{table}[t]
\Large
\centering
\resizebox{\linewidth}{!}{%
\begin{tabular}{ccccccc}
    \toprule
    Deterministic & ETH & HOTEL & UNIV & ZARA1 & ZARA2 & AVG \\ \midrule
    \STGCNN              & 1.04\pslp1.93 & 0.88\pslp1.63 & 0.71\pslp1.34 & 0.66\pslp1.22 & 0.57\pslp1.06 & 0.77\pslp1.44 \\
    \PECNet              & 1.20\pslp2.73 & 0.68\pslp1.51 & 0.78\pslp1.71 & 0.82\pslp1.85 & 0.62\pslp1.46 & 0.82\pslp1.85 \\
    \AgentFormer         & 1.67\pslp3.65 & 1.61\pslp3.59 & 1.62\pslp3.53 & 1.85\pslp4.13 & 1.68\pslp3.74 & 1.69\pslp3.73 \\
    \MID                 & 1.42\pslp2.94 & 0.64\pslp1.30 & 0.76\pslp1.62 & 0.74\pslp1.59 & 0.60\pslp1.31 & 0.83\pslp1.75 \\
    \ETFullCompact       & \tul{0.93}\pslp2.05 & 0.33\pslp0.64 & 0.58\pslp1.23 & 0.45\pslp0.99 & \tul{0.34}\pslp0.75 & 0.53\pslp1.13 \\ \cmidrule(lr){1-7}
    \SocialLSTM          & 1.09\pslp2.35 & 0.79\pslp1.76 & 0.67\pslp1.40 & 0.47\pslp1.00 & 0.56\pslp1.17 & 0.72\pslp1.54 \\
    \SocialGAN           & 1.13\pslp2.21 & 1.01\pslp2.18 & 0.60\pslp1.28 & \tbf{0.42}\pslp\tbf{0.91} & 0.52\pslp1.11 & 0.67\pslp1.41 \\
    \SRLSTM              & 1.01\pslp\tul{1.93} & 0.35\pslp0.72 & 0.66\pslp1.38 & 0.56\pslp1.23 & 0.44\pslp0.90 & 0.60\pslp1.23 \\
    \STARD               & 0.97\pslp2.00 & \tul{0.32}\pslp0.73 & \tul{0.56}\pslp1.25 & \tul{0.44}\pslp0.96 & 0.35\pslp0.77 & 0.53\pslp1.14 \\
    \Trajectronpp        & 1.02\pslp2.09 & 0.33\pslp\tul{0.63} & \tbf{0.52}\pslp\tul{1.16} & \tbf{0.42}\pslp0.94 & \tbf{0.32}\pslp\tbf{0.71} & \tul{0.52}\pslp\tul{1.11} \\ \cmidrule(lr){1-7}
    \tbf{\STFullCompact} & \tbf{0.72}\pslp\tbf{1.23} & \tbf{0.27}\pslp\tbf{0.50} & 0.57\pslp\tbf{1.12} & 0.44\pslp\tul{0.93} & 0.35\pslp\tul{0.73} & \tbf{0.47}\pslp\tbf{0.90} \\ 
    \bottomrule
\end{tabular}
}
\vspace{-3mm}
\caption{Evaluation on the deterministic trajectory prediction task.}
\label{tab:result_deterministic}
\end{table}

\subsection{Evaluation Results}\label{sec:experiment_result}
\vspace{0mm}\noindent\textbf{Stochastic prediction task.}\quad 
In~\cref{tab:result_stochastic}, since the ETH/UCY dataset release, many state-of-the-art models have made new records. Among them, our model also achieves performance competitive with the comparison methods. Our model regards historical, social, and environmental conditions of observations and images as the most important factors for successfully generating multimodal trajectories.

\begin{figure*}[t]
\centering
\includegraphics[width=\linewidth,trim={0 68mm 0 0},clip]{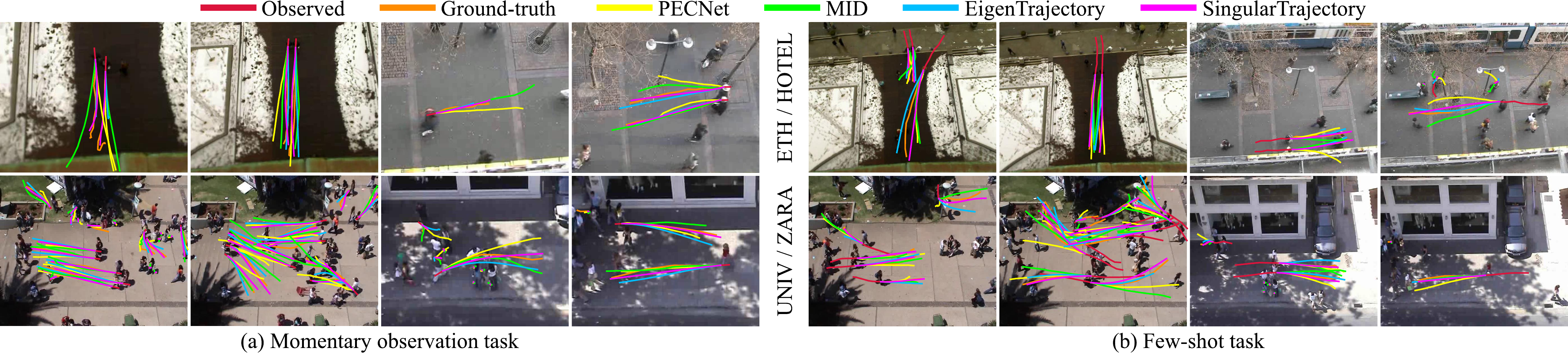}
\vspace{-7mm}
\caption{Visualization of prediction results on (a) momentary observation task and (b) few-shot task. To aid visualization, the best trajectory among $S\!=\!20$ samples are reported.}
\label{fig:result_qualitative}
\end{figure*}

\vspace{0.5mm}\noindent\textbf{Deterministic prediction task.}\quad 
Next, we evaluate our model in the deterministic trajectory prediction task in~\cref{tab:result_deterministic}. Our SingularTrajectory model has the best performance over both these latest state-of-the-art models, and the models optimized for deterministic prediction. In particular, our model exhibits significant performance improvements in the ETH and HOTEL scenes. These scenes have noisy observation paths, which have a negative impact on the prediction. We are able to achieve accurate predictions by leveraging the overall motion flow, which acts as a low-pass filter over the noisy sequence, in Singular space.

\vspace{0.5mm}\noindent\textbf{Momentary observation task.}\quad 
In~\cref{tab:result_momentary}, the state-of-the-art predictors for momentary observation mainly focus on the coordinates themselves, leading to a performance drop, given only two frames. Some models for this task try to bridge these gaps, but have not been sufficient. Fortunately, with the benefit of our Singular Space, even with only a two-frame input, our model can successfully represent the overall long-term flow. 
Note that our SingularTrajectory with two-frame observation demonstrates the closest performance to state-of-the-art stochastic prediction models with an entire history frame; even this is achieved without any masked trajectory complement or knowledge distillation.
Consequently, this allows it to accurately pinpoint the future locations of pedestrians, whose examples are displayed in~\cref{fig:result_qualitative}.

\begin{table}[t]
\Large
\centering
\resizebox{\linewidth}{!}{%
\begin{tabular}{ccccccc}
    \toprule
    Momentary & ETH & HOTEL & UNIV & ZARA1 & ZARA2 & AVG \\ \midrule
    \STGCNN              & 1.24\pslp2.23 & 0.77\pslp1.44 & 0.45\pslp0.81 & 0.38\pslp0.57 & 0.35\pslp0.58 & 0.64\pslp1.13 \\
    \PECNet              & 0.63\pslp1.04 & 0.28\pslp0.53 & 0.28\pslp0.49 & 0.25\pslp0.44 & 0.19\pslp0.34 & 0.33\pslp0.57 \\
    \AgentFormer         & 1.10\pslp2.11 & 0.50\pslp1.02 & 0.52\pslp1.10 & 0.56\pslp1.18 & 0.43\pslp0.89 & 0.62\pslp1.26 \\
    \MID                 & 0.63\pslp1.05 & 0.29\pslp0.49 & 0.30\pslp0.56 & 0.30\pslp0.56 & 0.22\pslp0.40 & 0.35\pslp0.61 \\
    \ETFullCompact       & \tul{0.46}\pslp\tul{0.76} & \tbf{0.17}\pslp\tbf{0.28} & \tul{0.25}\pslp\tul{0.44} & \tbf{0.19}\pslp\tul{0.35} & \tbf{0.15}\pslp\tbf{0.27} & \tbf{0.25}\pslp\tul{0.42} \\ \cmidrule(lr){1-7}
    \STT                 & 0.72\pslp1.45 & 0.48\pslp0.48 & 0.53\pslp1.09 & 0.64\pslp1.21 & 0.44\pslp0.88 & 0.57\pslp0.93 \\
    \STTDTO              & 0.62\pslp1.22 & 0.29\pslp0.56 & 0.58\pslp1.14 & 0.45\pslp0.98 & 0.34\pslp0.74 & 0.46\pslp0.93 \\
    \MOENext             & 0.71\pslp1.57 & 0.30\pslp0.58 & 0.52\pslp1.12 & 0.38\pslp0.81 & 0.33\pslp0.73 & 0.45\pslp0.96 \\
    \MOETrajectronpp     & 0.64\pslp1.12 & 0.20\pslp0.33 & 0.33\pslp0.62 & \tul{0.22}\pslp0.42 & \tul{0.17}\pslp0.32 & \tul{0.31}\pslp0.56 \\ \cmidrule(lr){1-7}
    \tbf{\STFullCompact} & \tbf{0.45}\pslp\tbf{0.67} & \tul{0.18}\pslp\tul{0.29} & \tbf{0.24}\pslp\tbf{0.43} & \tbf{0.19}\pslp\tbf{0.33} & \tul{0.17}\pslp\tul{0.28} & \tbf{0.25}\pslp\tbf{0.40} \\ 
    \bottomrule
\end{tabular}
}
\vspace{-3mm}
\caption{Evaluation on the momentary observation task.}
\label{tab:result_momentary}
\end{table}

\vspace{0.5mm}\noindent\textbf{Domain adaptation task.}\quad 
We next evaluate the performance of SingularTrajectory in a domain adaptation task. For simplicity, the ETH, HOTEL, UNIV, ZARA1, and ZARA2 scenes are denoted as A, B, C, D, and E, respectively. For example, `A2B' means that a model is trained on the ETH scene and tested on the HOTEL scene. As demonstrated in~\cref{tab:result_domain}, our SingularTrajectory model shows the performance nearly equivalent to those of models specifically designed for deterministic prediction. Particularly, our model produces impressive results in the challenging B2A, C2A, D2A and E2A scenarios where even models specialized in domain adaptation fail. Our SingularTrajectory framework specializes in learning general human motions, and is not limited to a specific domain, enabling accurate predictions even in extreme cases.

\begin{figure}[t]
\centering
\includegraphics[width=\linewidth,trim={0 68mm 366.8mm 0},clip]{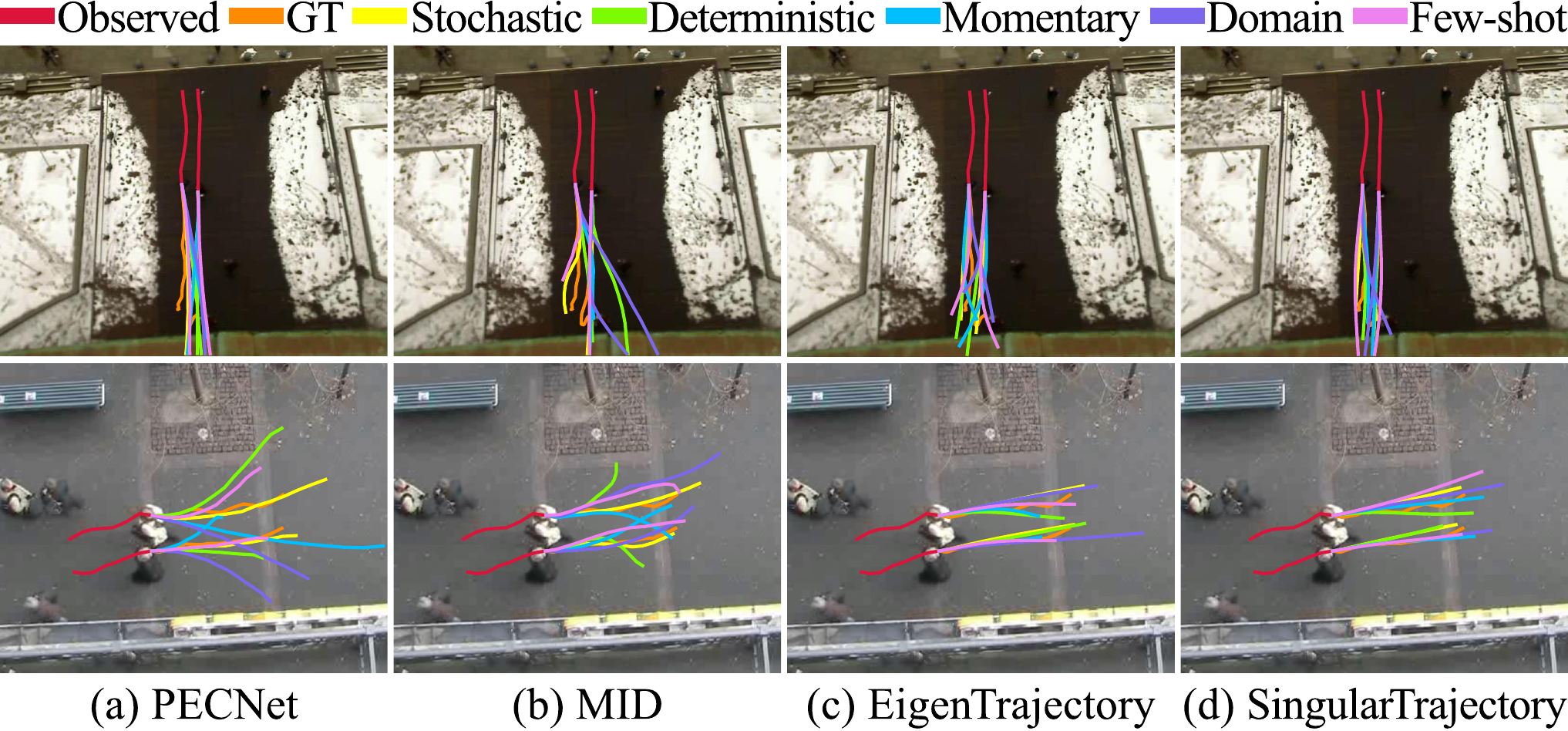}
\vspace{-7mm}
\caption{Visualization of prediction consistency across five tasks. The more consistent the prediction is the better.}
\label{fig:result_differences}
\end{figure}

\vspace{0.5mm}\noindent\textbf{Few-shot task.}\quad 
Finally, the results for the few-shot task are reported in~\cref{tab:result_fewshot}, whose examples are in~\cref{fig:result_qualitative}. As expected, our SingularTrajectory significantly outperforms the existing models. With limited data, the existing works, except the diffusion-based model, tend to easily overfit. In contrast, our methods make it possible to explicitly designate future prototype paths, while taking full advantage of the expression ability of the diffusion model. In particular, even with only 10\% of the data, our model achieves substantial improvements with respect to both data efficiency and performance. 
\Cref{fig:result_differences} illustrates several cases where there are differences between the predictions of SingularTrajectory and other comparison methods. Previous works often show significant performance drops in other associated tasks compared to stochastic prediction tasks. In contrast, our model consistently predicts the best trajectories across multiple tasks.

\begin{table}[t]
\Large
\centering
\resizebox{\linewidth}{!}{%
\begin{tabular}{ccccccc}
    \toprule
    Few-Shot & ETH & HOTEL & UNIV & ZARA1 & ZARA2 & AVG \\ \midrule
    \STGCNN              & 0.89\pslp1.63 & 1.22\pslp2.48 & 0.90\pslp1.61 & 0.68\pslp1.25 & 1.36\pslp2.12 & 1.01\pslp1.82 \\
    \PECNet              & 0.72\pslp1.46 & 0.29\pslp0.53 & 0.58\pslp0.93 & \tul{0.27}\pslp0.44 & 0.22\pslp0.38 & 0.41\pslp0.75 \\
    \AgentFormer         & 1.60\pslp2.65 & 1.02\pslp1.64 & 1.13\pslp1.90 & 1.19\pslp2.01 & 1.08\pslp1.59 & 1.20\pslp1.96 \\
    \MID                 & 0.57\pslp0.92 & 0.21\pslp0.33 & 0.32\pslp0.60 & \tul{0.27}\pslp0.49 & 0.24\pslp0.42 & 0.32\pslp0.55 \\
    \ETFullCompact       & 0.39\pslp0.64 & 0.13\pslp0.21 & \tbf{0.25}\pslp\tbf{0.43} & \tbf{0.21}\pslp\tul{0.39} & \tbf{0.15}\pslp\tbf{0.27} & \tul{0.23}\pslp\tul{0.39} \\ \cmidrule(lr){1-7}
    \tbf{\STFullCompact} & \tbf{0.35}\pslp\tbf{0.46} & \tbf{0.14}\pslp\tbf{0.21} & \tbf{0.26}\pslp\tbf{0.44} & \tbf{0.21}\pslp\tbf{0.36} & \tul{0.18}\pslp\tul{0.31} & \tbf{0.23}\pslp\tbf{0.35} \\ 
    \bottomrule
\end{tabular}
}
\vspace{-3mm}
\caption{Evaluation on the few-shot trajectory prediction task.}
\label{tab:result_fewshot}
\end{table}

\subsection{Ablation Studies}\label{sec:experiment_ablation} 
\vspace{0mm}\noindent\textbf{The number of motion vectors $\bm{K}$.}\quad
First, we conduct a component study by varying the dimension $K$ of Singular space in~\cref{tab:ablation_dimension}. To find the best number of singular vectors in general, we carry out experiments across all five tasks using ZARA1 scene where there are both human-environmental and human-human interactions, following~\cite{xu2022tgnn}. As the number of motion vectors $K$ increases, more detailed movements are captured. In contrast, when using the smaller $K$, it compresses the space, mainly to represent the overall motion flow. The performance tends to plateau when $K$ is larger than $3$, as long as $K$ is not too small to cover most movements. We set $K\!=\!4$ as the dimension for the Singular space because it shows the most effective prediction results across all tasks.

\begin{table}[t]
\Large
\centering
\resizebox{\linewidth}{!}{%
\begin{tabular}{cccccccc}
    \toprule
    ~~~~~~$K$~~~~~~ & \!\!\!\!\!\!\!Deterministic\!\!\!\!\!\!\! & \!\!Stochastic\!\! & \!\!\!\!Momentary\!\!\!\! & Domain & Few-shot & Average \\ \midrule
    1 & 0.52 / 1.03 & 0.30 / 0.55 & 0.31 / 0.58 & 0.51 / 1.07 & 0.34 / 0.61 & 0.40 / 0.77 \\
    2 & 0.51 / 1.01 & \tul{0.20} / \tbf{0.32} & \tul{0.20} / \tbf{0.33} & 0.50 / 1.06 & 0.23 / \tul{0.37} & 0.33 / 0.62 \\
    3 & 0.45 / \tbf{0.93} & \tbf{0.19} / \tul{0.33} & \tul{0.20} / \tbf{0.33} & \tul{0.47} / \tbf{1.03} & \tul{0.22} / 0.38 & \tul{0.31} / \tul{0.60} \\
    4 & \tul{0.44} / \tbf{0.93} & \tbf{0.19} / \tbf{0.32} & \tbf{0.19} / \tbf{0.33} & \tul{0.47} / \tbf{1.03} & \tbf{0.21} / \tbf{0.36} & \tbf{0.30} / \tbf{0.59} \\
    5 & \tbf{0.43} / \tul{0.94} & \tbf{0.19} / \tul{0.33} & \tul{0.20} / \tul{0.34} & \tbf{0.46} / \tul{1.04} & \tul{0.22} / \tul{0.37} & \tbf{0.30} / \tul{0.60} \\
    6 & \tbf{0.43} / \tul{0.94} & \tbf{0.19} / \tbf{0.32} & \tul{0.20} / \tul{0.34} & \tbf{0.46} / \tul{1.04} & \tul{0.22} / \tul{0.37} & \tbf{0.30} / \tul{0.60} \\ 
    \bottomrule
\end{tabular}
}
\vspace{-3mm}
\caption{Ablation study on the Singular space dimension $K$.}
\label{tab:ablation_dimension}
\end{table}

\begin{table}[t]
\Large
\centering
\resizebox{\linewidth}{!}{%
\begin{tabular}{ccccccc}
    \toprule
    ~~Adoption~~ & \!\!\!\!\!\!\!Deterministic\!\!\!\!\!\!\! & \!\!Stochastic\!\! & \!\!\!\!Momentary\!\!\!\! & Domain & Few-shot & Average \\ \midrule
    Direct      & 0.75 / 1.47 & 0.27 / 0.48 & 0.31 / 0.54 & 0.78 / 1.60 & 0.27 / 0.48 & 0.48 / 0.91 \\
    Initial     & \tul{0.47} / \tul{0.95} & \tul{0.22} / \tul{0.40} & \tul{0.22} / \tul{0.39} & \tul{0.50} / \tul{1.17} & \tul{0.23} / \tul{0.40} & \tul{0.33} / \tul{0.66} \\
    Residual    & \tbf{0.44} / \tbf{0.93} & \tbf{0.19} / \tbf{0.32} & \tbf{0.19} / \tbf{0.33} & \tbf{0.47} / \tbf{1.03} & \tbf{0.21} / \tbf{0.36} & \tbf{0.30} / \tbf{0.59} \\ \cmidrule(lr){1-7}
    Independent & 0.46 / 0.94 & 0.21 / 0.39 & 0.21 / 0.39 & 0.49 / 1.16 & 0.21 / 0.39 & 0.32 / 0.65 \\
    Jointly     & \tbf{0.44} / \tbf{0.93} & \tbf{0.19} / \tbf{0.32} & \tbf{0.19} / \tbf{0.33} & \tbf{0.47} / \tbf{1.03} & \tbf{0.21} / \tbf{0.36} & \tbf{0.30} / \tbf{0.59} \\
    \bottomrule
\end{tabular}
}
\vspace{-3mm}
\caption{Ablation study on the adoption of diffusion model.}
\label{tab:ablation_type}
\end{table}

\begin{figure}[t]
\centering
\includegraphics[width=\linewidth,trim={0 52mm 0 0},clip]{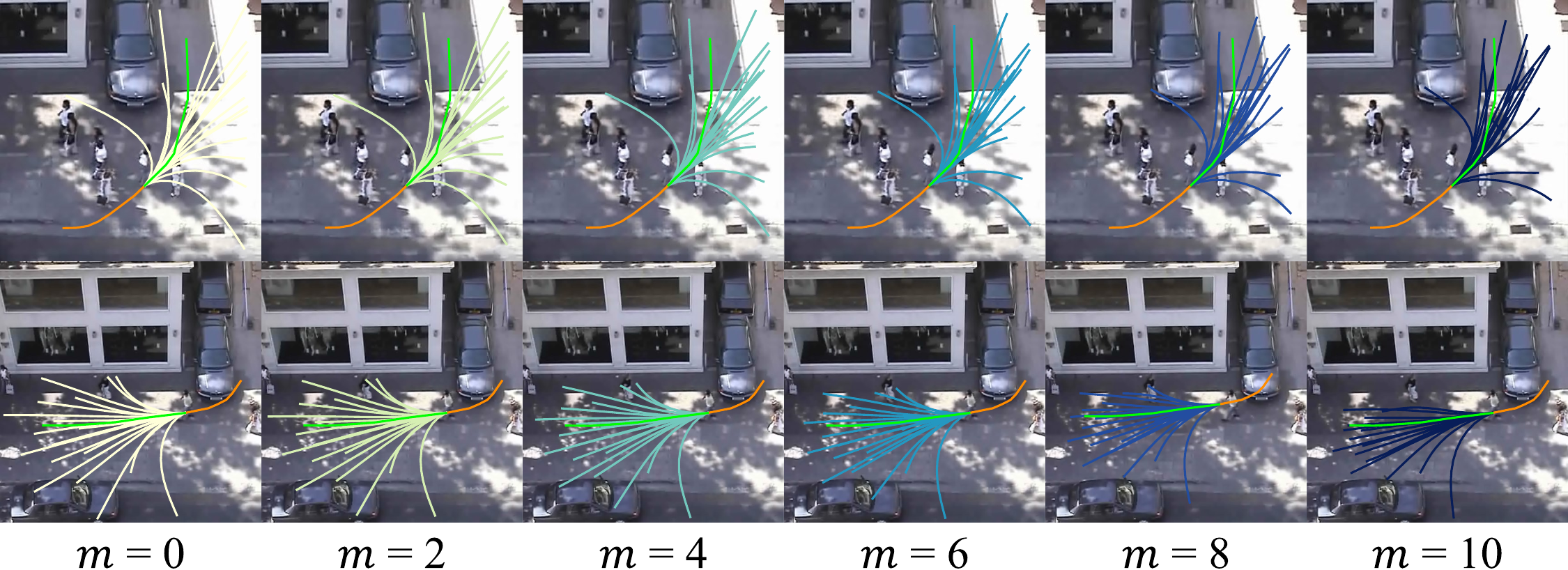}
\vspace{-7mm}
\caption{Visualization\,of\,anchor\,refinement.\,The\,denoising\,process progressively refines the prototype paths at each diffusion step $m$.}
\label{fig:result_denoising}
\end{figure}

\vspace{0.5mm}\noindent\textbf{Trajectory denoising methods.}\quad 
Next, we evaluate three types of diffusion models for the trajectory prediction tasks, as shown in~\cref{tab:ablation_type}. First, we use a basic model similar to MID~\cite{gu2022mid}, which directly denoises from a Gaussian noise to a trajectory. This method fails to achieve good performances. The use of an anchor as an intermediate state, similar to LED~\cite{mao2023leapfrog}, which reduces the denoising steps by skipping the initial denoising steps, seems to validate its generality. This demonstrates that our adaptive anchor can function as a good initializer for the diffusion model, particularly showing nearly identical outcomes in the stochastic, momentary observation, and few-shot tasks when predicting a multimodal path. However, we confirm that our scheme, which denoises only a residual by adding perturbation to a prototype path in~\cref{fig:result_denoising}, showcases the best performance in all the tasks. Additionally, compared to refining the prototype paths independently, regarding them as a batch dimension, our model can accurately predict the future when prototype paths are jointly refined.

\vspace{0.5mm}\noindent\textbf{Diffusion steps $\bm{M}$.}\quad 
Lastly, we check how many steps in the diffusion model are needed for the cascaded refinement of the adaptive anchor. In~\cref{tab:ablation_step}, we confirm that the best performance comes from $M\!=\!10$. This is because the DDIM scheduler accelerates its convergence and the prototype path provides a rough initial trajectory, and so fewer denoising steps are sufficient. However, as the number of denoising steps increases, the information in the initial prototype path becomes attenuated due to the noise. As a result, we observe a slight decrease in performance.

\section{Conclusion}
In this study, we introduce SingularTrajectory, a universal trajectory predictor model for all related trajectory prediction tasks. By unifying trajectory modalities into one Singular space, our approach standardizes trajectory data with shared motion dynamics, which eliminates the need for task-specific adjustments. The incorporation of an adaptive anchor system further personalizes the prototype paths, allowing them to interpret and adapt to environmental contexts and enhancing the reliability of the trajectory prediction. By successfully incorporating Singular space into the diffusion model, our SingularTrajectory framework successfully achieves the state-of-the-art results across five different benchmarks. This establishes SingularTrajectory itself as a general solution that covers a multitude of scenarios.

\vspace{1mm}
\fontsize{8.6}{10.9}\selectfont{%
\noindent\textbf{Acknowledgement}
This research was supported by 'Project for Science and Technology Opens the Future of the Region' program through the INNOPOLIS FOUNDATION funded by Ministry of Science and ICT (Project Number: 2022-DD-UP-0312), Vehicles AI Convergence Research $\&$ Development Program through the National IT Industry Promotion Agency of Korea (NIPA) funded by the Ministry of Science and ICT (No.S1602-20-1001), and the Institute of Information $\&$ communications Technology Planning $\&$ Evaluation (IITP) grant funded by the Korea government (MSIT) (No.2019-0-01842, Artificial Intelligence Graduate School Program (GIST).}

\begin{table}[t]
\Large
\centering
\resizebox{\linewidth}{!}{%
\begin{tabular}{cccccccc}
    \toprule
    ~~~~~~$M$~~~~~~ & \!\!\!\!\!\!\!Deterministic\!\!\!\!\!\!\! & \!\!Stochastic\!\! & \!\!\!\!Momentary\!\!\!\! & Domain & Few-shot & Average \\ \midrule
    1   & 0.45 / 0.95 & 0.20 / 0.34       & \tul{0.20} / 0.35       & 0.48 / 1.05       & 0.21 / 0.37 & 0.31 / 0.61 \\
    2   & 0.44 / 0.93 & 0.19 / \tul{0.33} & \tul{0.20} / \tul{0.34} & 0.47 / \tul{1.04} & 0.20 / 0.36 & 0.30 / \tul{0.60} \\
    5   & 0.44 / 0.93 & 0.19 / \tul{0.33} & \tul{0.20} / \tul{0.34} & 0.47 / \tul{1.04} & 0.20 / 0.36 & 0.30 / \tul{0.60} \\
    10  & 0.44 / 0.93 & 0.19 / \tbf{0.32} & \tbf{0.19} / \tbf{0.33} & 0.47 / \tbf{1.03} & 0.21 / 0.36 & 0.30 / \tbf{0.59} \\
    25  & 0.44 / 0.93 & 0.19 / \tbf{0.32} & \tul{0.20} / \tul{0.34} & 0.47 / \tbf{1.03} & 0.20 / 0.36 & 0.30 / \tul{0.60} \\
    \bottomrule
\end{tabular}
}
\vspace{-3mm}
\caption{Ablation study on the diffusion steps $M$.}
\label{tab:ablation_step}
\end{table}

{
    \small
    \bibliographystyle{ieeenat_fullname}
    \bibliography{egbib}
}

\end{document}